\documentclass[10pt,twocolumn,letterpaper]{article}

\usepackage[pagenumbers]{cvpr} % To force page numbers, e.g. for an arXiv version

\definecolor{green}{RGB}{0,150,10}

\newcommand{\mysection}[1]{\noindent\textbf{#1.}}

\newlength\mytmplen

\definecolor{cvprblue}{rgb}{0.21,0.49,0.74}
\usepackage[pagebackref,breaklinks,colorlinks,allcolors=cvprblue]{hyperref}
\usepackage{algorithm}
\usepackage{algpseudocode}
\usepackage{multirow}
\usepackage{soul} % Make sure to include this in the preamble

\def\paperID{****} % *** Enter the Paper ID here
\def\confName{CVPR}
\def\confYear{2025}
\newcommand{\methodname}{ZeroKey\xspace}

\title{\methodname: Point-Level Reasoning and Zero-Shot 3D Keypoint Detection from Large Language Models}

\author{
\textbf{Bingchen Gong$^{1}$} \quad \quad
\textbf{Diego Gomez$^{1}$} \quad \quad
\textbf{Abdullah Hamdi$^{2}$}\quad \quad
\textbf{Abdelrahman Eldesokey$^{3}$}\\
\textbf{Ahmed Abdelreheem$^{3}$}\quad \quad
\textbf{Peter Wonka$^{3}$}\quad \quad
\textbf{Maks Ovsjanikov$^{1}$}
\\ \\
 $^1$École Polytechnique \quad \quad \quad $^2$Visual Geometry Group, University of Oxford\\
 $^3$King Abdullah University of Science and Technology (KAUST) \\ \small
\texttt{gongbingchen@gmail.com}
}

\begin{document}

\twocolumn[{%
\renewcommand\twocolumn[1][]{#1}%
\maketitle
\vspace{-2em}
\includegraphics[width=\linewidth,trim={0 0 0 0},clip]{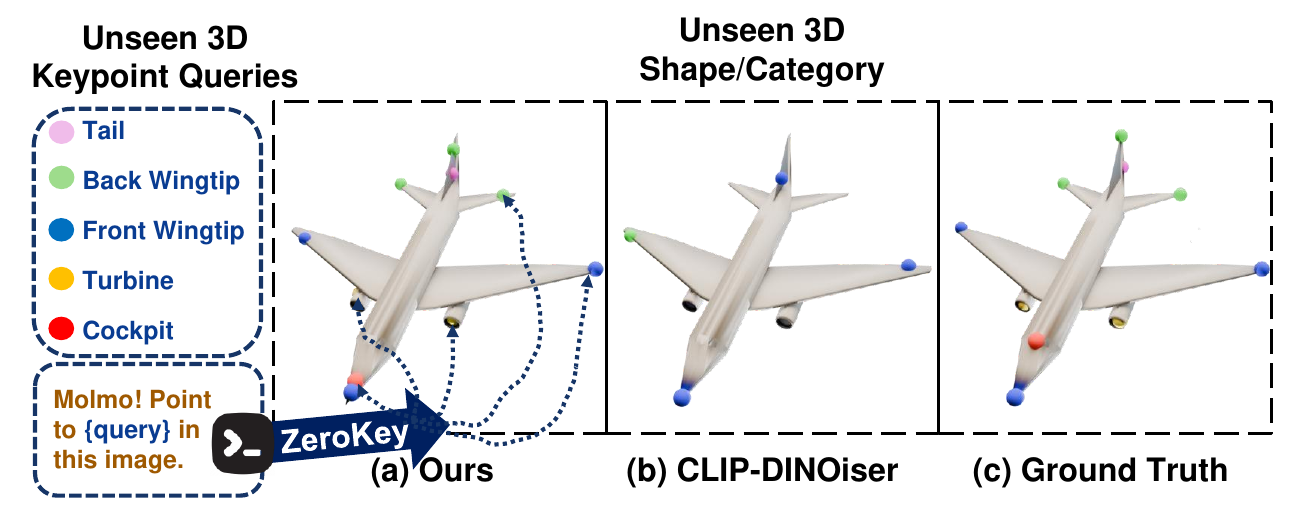}
\vspace{-8mm}
\captionof{figure}{\textbf{Zero-shot 3D Keypoint Detection.} Without any ground truth labels or supervised training, our method leverages the point-level reasoning embedded within MLLMs to extract and name salient keypoints on 3D models. The figure illustrates how our approach achieves competitive performance compared to CLIP-DINOiser \cite{wysoczanska2024clip} baselines, highlighting the potential of integrating language models with vision tasks for enhanced 3D shape understanding. \vspace{1.0em}}
\label{fig:teaser}
}]

\begin{abstract}
We propose a novel zero-shot approach for keypoint detection on 3D shapes.
Point-level reasoning on visual data is challenging as it requires precise localization capability, posing problems even for powerful models like DINO or CLIP.
Traditional methods for 3D keypoint detection rely heavily on annotated 3D datasets and extensive supervised training, limiting their scalability and applicability to new categories or domains. In contrast, our method utilizes the rich knowledge embedded within Multi-Modal Large Language Models (MLLMs). Specifically, we demonstrate, for the first time, that pixel-level annotations used to train recent MLLMs can be exploited for both extracting and naming salient keypoints on 3D models without any ground truth labels or supervision. 
Experimental evaluations demonstrate that our approach achieves competitive performance on standard benchmarks compared to supervised methods, despite not requiring any 3D keypoint annotations during training. Our results highlight the potential of integrating language models for localized 3D shape understanding. This work opens new avenues for cross-modal learning and underscores the effectiveness of MLLMs in contributing to 3D computer vision challenges. 
Our code is available at \url{https://sites.google.com/view/zerokey}.
\end{abstract}    

\vspace{-5mm}
\section{Introduction}
Multimodal Large Language Models (MLLMs) have been shown to seamlessly integrate visual and text representations with great success \cite{GPT4,zhu2023minigpt,liu2023llava,liu2023improvedllava,alayrac2022flamingo,li2023blip}. Models like GPT-4 with vision capabilities, OFA~\cite{wang2022ofa}, and Flamingo~\cite{alayrac2022flamingo} have demonstrated the ability to process visual and textual data, leading to advancements in vision-language understanding. Furthermore, related efforts have been made to endow MLLMs with 3D or spatial reasoning \cite{hong20233d,zhou2023uni3d,xu2024pointllm}. When trained at scale, these MLLMs enable tackling a wide array of complex tasks, ranging from comprehensive image or 3D shape description to answering detailed questions about visual content. 

Unfortunately, despite this tremendous progress, existing MLLMs still struggle with tasks that require precise points or pixel-level reasoning, such as localization, counting, or salient keypoint detection \cite{Zhang_2024_CVPR, rahmanzadehgervi2024vision, campbell2024understanding}. This type of reasoning focuses on understanding and interpreting visual input at a fine-grained level using text. In general, we can observe an increased level of difficulty when going from complete objects to object parts and, finally, to specific \textit{points} or small regions. Even advanced models like GPT-4o \cite{GPT4} and Claude Sonnet 3.5 \cite{anthropic2024claude}, which have demonstrated impressive capabilities in various computer vision tasks, still struggle with point-level understanding.

This challenge arises because point-level reasoning requires models to accurately identify and analyze local details, such as landmarks on a face, joints in a human pose, or intricate components in a 3D shape. However, the architectures, training methods, and dataset annotations of most existing approaches are primarily designed to capture global visual properties and potentially link them to textual descriptions. Since text captions are most often crawled from the Internet, and thus rarely contain precise point or region-level annotations, point-level reasoning capabilities of most MLLMs remain very limited.

Addressing this problem necessitates a shift in how models are designed and trained. One potential solution is to use more powerful visual encoders trained on real-world data encompassing a variety of global and local features. Interestingly, scaling alone to more crawled data or model complexity does not seem to alleviate this problem \cite{peng2024synthesize}. In contrast, very recently, a family of MLLMs has been introduced that was trained with specially-designed auxiliary training data and tasks aimed at pixel-level image annotations \textit{alongside} global image captions \cite{deitke2024molmo}. When trained with this additional data source, the resulting model (dubbed Molmo) exhibits impressive localized reasoning abilities (counting, localization, etc.) surpassing other MLLMs.

Inspired by these recent developments, we propose investigating MLLMs endowed with point-level reasoning in the context of 3D shape understanding and specifically for zero-shot keypoint detection. Given a 3D model from an arbitrary category, our main task is to localize and name salient keypoints on this model. Traditional methods for 3D keypoint detection heavily rely on annotated 3D datasets and extensive supervised training \cite{wimmer2024back, attaiki2022ncp, you2022ukpgan}.  This severely limits their applicability to specific data-rich object categories. At the same time, recent general-purpose 3D LLMs are typically trained with explicit objectives that align 3D data representations with vision-language models \cite{hong20233d}, enabling global tasks like classification. However, since they are based on standard vision-language models with limited localization capabilities, none of these methods can accurately perform 3D keypoint localization based on a language prompt, see the supplementary material for more details.

Even in the presence of a powerful MLLM, we note that localization or keypoint detection from images to 3D environments presents a significant challenge. Since both detections and point-level captions tend to be noisy, developing a robust approach requires reliable multi-view aggregation, backprojection, and filtering mechanisms. In this paper, we propose a comprehensive zero-shot 3D keypoint detection method that exploits the point-level reasoning capabilities of recent MLLMs \cite{deitke2024molmo} while integrating 3D consistency and being completely category agnostic. To the best of our knowledge, ours is the first robust method for zero-shot 3D keypoint detection.

The applications of this approach are vast. This paper demonstrates two significant applications: Schelling points analysis and point describability and consistency. 
Our evaluations demonstrate that, despite not having access to ground truth data, our method can predict points similar to those identified by human annotators by selecting appropriate text prompts. Additionally, we developed two text-image baselines for detecting 3D keypoints from text prompts and conducted comparisons with these baselines.

To summarize, our main contributions are as follows:
\begin{itemize}
\item Identify the challenge of zero-shot point-level reasoning in 3D computer vision.
\item Propose the first zero-shot 3D keypoint detection method that operates without 3D annotations or training and applies to arbitrary shape categories, establishing a baseline for zero-shot 3D keypoint detection in this new area.
\item Evaluate several MLLMs in our context, highlighting the importance of auxiliary point-level training to solve the task at hand.
\end{itemize}

\section{Related Work}

\begin{figure*}[t]
  \centering
   \includegraphics[width=0.9\linewidth]{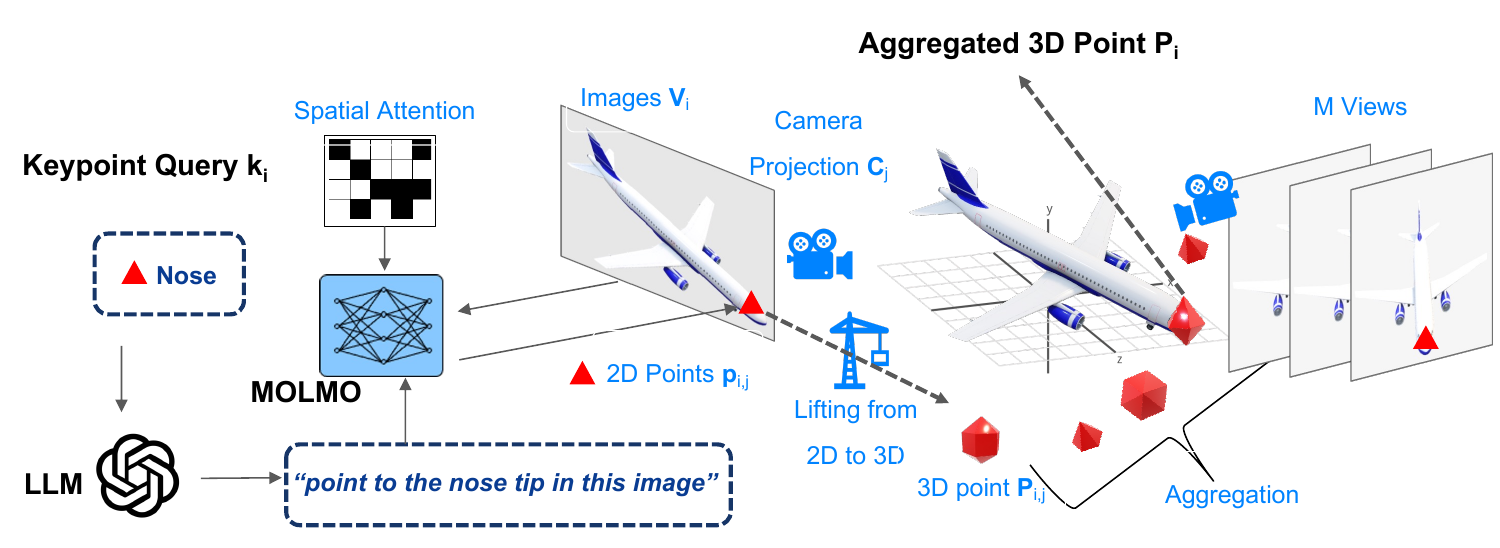}
\caption{\textbf{\methodname Pipeline.} Our proposed \methodname employs MLLM Molmo for zero-shot keypoint detection on 3D objects by 1) rendering multiple views for a given shape, 2) leveraging MLLM reasoning in each view using point-specific prompts, and 3) aggregating the results through clustering, eliminating the need for annotated training data for 3D keypoint detection. }
\vspace{-1em}
\label{fig:pipeline}
\end{figure*}

\subsection{Multimodal Large Language Models}

The remarkable success of vision-language models like CLIP \citep{CLIP} and large language models like GPT-3 \citep{brown2020language} has motivated the integration of multimodal capabilities into large language models for visual understanding. Multimodal Large Language Models (MLLMs) have been extended to process images alongside text, achieving remarkable success in vision-language tasks \citep{GPT4,zhu2023minigpt,liu2023llava,liu2023improvedllava,alayrac2022flamingo,li2023blip}. Notable models such as GPT-4 with vision capabilities, MiniGPT-4 \citep{zhu2023minigpt}, LLaVA \citep{liu2023llava,liu2023improvedllava}, Flamingo \citep{alayrac2022flamingo}, and BLIP \citep{li2023blip} have demonstrated advanced abilities in integrating visual and textual information, enabling tasks ranging from detailed image captioning to complex visual reasoning. These models leverage large-scale pre-training on extensive image-text datasets to capture both global semantics and fine-grained details in 2D images. Despite their success with 2D data, extending these capabilities to 3D data remains a significant challenge due to the elevated cost and complexity of collecting large amounts of high-quality 3D data. Consequently, there is no dominant MLLM capable of processing 3D data with natural language understanding. To address this gap, recent efforts have focused on aligning models trained on small amounts of 3D data with large vision-language models. For instance, ULIP \citep{xue2023ulip} seeks to align 3D models with vision-language representations to enhance cross-modal understanding. Similarly, Hong et al. \citep{hong20233d} propose training a 3D Large Language Model by leveraging a pre-trained vision-language model, aiming to bridge the modality gap and facilitate better 3D comprehension through multimodal learning. While these ``3D LLMs" are promising, the granularity we are targeting in work at the point-level 3D reasoning is still beyond their current capabilities \cite{hong20233d,xue2023ulip}.

\subsection{Keypoint Detection}
Keypoint detection has been extensively studied in 3D due to its significance in shape representation, matching, and abstraction applications \cite{jakab2021keypointdeformer,wang2018learning,shi2021skeleton,suwajanakorn2018discovery,fernandez2020unsupervised,chen2020unsupervised,yew20183dfeat}. 
However, the definition of a keypoint has remained task-specific and subjective. 
Even human annotations of keypoints often differ, as highlighted in \cite{you2020keypointnet,chen2012schelling}. 
Recent work \cite{hedlin2024unsupervised} has shown that image diffusion models can develop an unsupervised understanding of keypoints as salient features. 
Additionally, it was demonstrated in \cite{shtedritski2023does} that the CLIP foundation model \cite{CLIP} can reasonably interpret keypoints with appropriate visual prompts.
This paper investigates the understanding of keypoints as foundational elements of objects and shapes within recent multimodal large language models (MLLMs).

\subsection{Lifting from 2D to 3D}
A number of studies have explored multi-view approaches to enhance 3D data analysis by leveraging 2D representations \citep{mvsceneseg,mvvirtualsceneseg,mvlabeldifusion,mvshapeseg,mvpnet,distill3d,mvtn,Hamdi2024ijcv,mai2023egoloc,tracknerf}. A key challenge in this area is combining features from different viewpoints to represent local 3D points or voxels while preserving essential geometric details. Some methods address this by averaging features across views \citep{mvvirtualsceneseg,mvshapeseg}, propagating labels between views \citep{mvlabeldifusion}, or learning from reconstructed neighboring points \citep{mvpnet}. Others organize points within a unified grid structure \citep{learn2dfor3d} or merge multi-view features with 3D voxel information \citep{mvsceneseg,mvinsseg}. VointNet \citep{hamdi2023voint} operates within the Voint cloud framework, retaining the original point cloud's compactness and descriptive 3D properties. It employs learned aggregation of multi-view features applied independently to each point. Recently, there's a growing trend of utilizing foundational models with zero-shot understanding capabilities—such as the Segment Anything Model \citep{kirillov2023segment}—to achieve zero-shot 3D comprehension by lifting 2D predictions into 3D space \citep{Huang2023Segment3D,Schult23ICRA,abdelreheem2023zero,abdelreheem2023satr}. Our work differs from the Zero-shot 3D segmentation in that our final goal is a fine-grained \textit{single} 3D point representing an unseen text query on an unseen new 3D object.  As we show below, existing vision-language models struggle with this task, thus necessitating a novel adapted solution.

\section{Motivation}

% Current problem why it fails
Localization and naming of points in an image or a 3D shape is an extremely challenging problem. Solving such a problem requires a localized understanding of the visual data and tight integration between text. Vision-Language Models (VLMs) are thus a strong candidate to tackle this problem. However, as mentioned above, most such models are trained using global or object-level semantics which are insufficient for point localization tasks. Furthermore, the recent MLLMs that incorporate 3D data \cite{hong20233d,zhou2023uni3d,xu2024pointllm} are typically trained with explicit alignment against pre-trained traditional vision-language models, and thus inherit their limitations in point-level reasoning. As a result, such models are unable to solve the zero-shot keypoint detection problem, which is the main focus of our work.

Existing methods for the 3D keypoint detection problem typically formulate the problem as either a supervised learning task, by exploiting the ground truth annotations, e.g., in the KeypointNet dataset \cite{you2020keypointnet}. Alternatively, several few-shot approaches have been developed, e.g., \cite{wimmer2024back}, which are based on \textit{transferring} known keypoints to new instances. 

Rather than training an entirely new 3D model, our goal is to exploit the improved point-level reasoning capabilities of very recent VLMs and demonstrate their utility in 3D shape understanding tasks. 
We base our approach on the recent open-source MOLMO model \cite{deitke2024molmo}, which incorporated a specially designed pixel-level annotation task into its training. Thanks to a well-chosen design and this additional curated dataset the model has been shown to perform significantly better than even the largest closed-source models in localized image understanding.

Our key motivation is to evaluate the effectiveness of such a model in the context of \textit{3D shape} understanding and specifically for zero-shot keypoint detection. We first introduce the problem of zero-shot keypoint detection and naming. Then, we develop a pipeline based on prompting the model and multi-view aggregation. We show that the resulting approach leads to unprecedented localized 3D shape understanding, opening the door to a range of applications.  
In this paper, we present the very first baseline for solving the zero-shot 3D keypoint detection problem. We leverage the powerful 2D spatially-aware VLM, Molmo \cite{deitke2024molmo}, to generate precise 2D keypoint predictions from several points of view. These points are back-projected into a 3D shape, and after aggregation, we obtain a 3D keypoint prediction.

\begin{figure}[t]
	\centering
	\includegraphics[width=\linewidth,trim={10 10 10 0},clip]{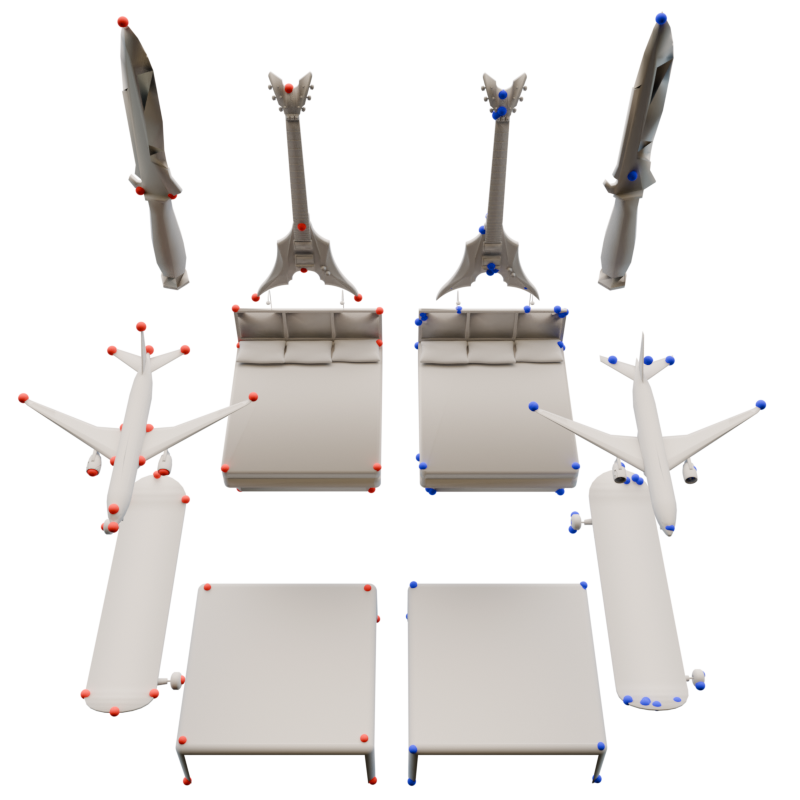}
	\caption{Comparing the ground truth KeypointNet dataset annotations (in red) to our method's predictions (in blue). This figure showcases our results on the KeypointNet dataset, illustrating the effectiveness of our approach in keypoint detection. The close alignment of the red and blue dots demonstrates the effectiveness of our approach in accurately detecting keypoints and highlights its precision in point-level reasoning.}
    \vspace{-1em}
	\label{fig:all_shapes}
\end{figure}

\section{Method}

\paragraph{Problem setting and overview:} We consider the Zero-Shot Keypoint Detection problem for 3D shapes. Specifically, given a 3D shape, we aim to automatically generate a set of salient points corresponding to the shape's semantic parts. 
%To the best of our knowledge, ours is the first zero-shot method that doesn't require any reference shape for correspondence mapping or labels from a dataset for training. 
Our solution comprises three main components: first, we prompt a MLLM with the shape, asking the model to generate a list of names for possible candidate keypoints. Then, for each candidate, we ask the model to detect the precise coordinates of the point in a given image. Finally, we back-project those detected points into 3D and aggregate them to get the location of 3D keypoints in a given shape.

\subsection{Generating Text Candidates for Salient Points}
To detect points using a MLLM, we first need to assign names to these points. The definition of keypoints can vary depending on the application. However, these points are typically common to a given shape, salient, and carry significant semantic meaning. This information can be retrieved from MLLMs by prompting the model with the shape class in text form or by using an image of the shape when the MLLM supports multimodal input.
Given the shape class label $c$, we prompt the MLLM to generate a list of candidate keypoint names:
$
\mathcal{K} = \{k_1, k_2, \dots, k_N\},
$
where $N$ is the number of keypoints generated. For example, for the class ``airplane,'' the MLLM might generate keypoint names such as ``nose,'' ``wing tip,'' and ``tail.''
In our experiments, we utilize GPT-4o \cite{GPT4} as our MLLM backbone for this step. We prompt it with a front-facing rendering of the 3D model along with the request text: \textit{``List possible salient key points (in text).''} The GPT-4o model is instructed to provide a candidate list of possible names for keypoints in JSON format. %For each 3D model, GPT-4o typically generates between 6 and 10 possible keypoint names.

\subsection{Prompting Molmo to Detect 2D Keypoints}
To detect the precise 2D coordinates of each candidate keypoint, we utilize Molmo~\cite{deitke2024molmo}, a state-of-the-art MLLM capable of localizing points in images based on natural language prompts.
Given a set of candidate keypoint names $\mathcal{K} = \{k_1, k_2, \dots, k_N\}$ and a set of images (views) of the 3D shape $\mathcal{V} = \{\mathbf{V}_1, \mathbf{V}_2, \dots, \mathbf{V}_M\}$, we prompt Molmo to detect the location of each keypoint in each image.
We define the detection function as:
\begin{equation}
\mathbf{p}_{i,j} = \text{Molmo}(\mathbf{V}_j, k_i), i = 1, \dots, N; j = 1, \dots, M,
\end{equation}
where $\mathbf{p}_{i,j} \in \mathbb{R}^2$ are the 2D coordinates of keypoint $k_i$ in image $\mathbf{V}_j$ as predicted by Molmo.
The prompt to Molmo consists of the image $\mathbf{V}_j$ and the instruction to localize the keypoint $k_i$. For example:
\begin{center}
    \textit{``Point to the left wing tip in this image.''}
\end{center}
This leverages Molmo's capability to understand natural language instructions and perform point-level localization. 
%\A{I think we should delve into exactly how Molmo does it mathematically, representing spatial attention, etc. Currently, this is a very abstract black-box definition of Molmo}

\subsection{Zero-Shot 3D Keypoint Detection}
\begin{figure}[t]
	\centering
	\begin{subfigure}[t]{0.32\linewidth}
		\includegraphics[width=\textwidth,trim={0 10 0 15},clip]
		{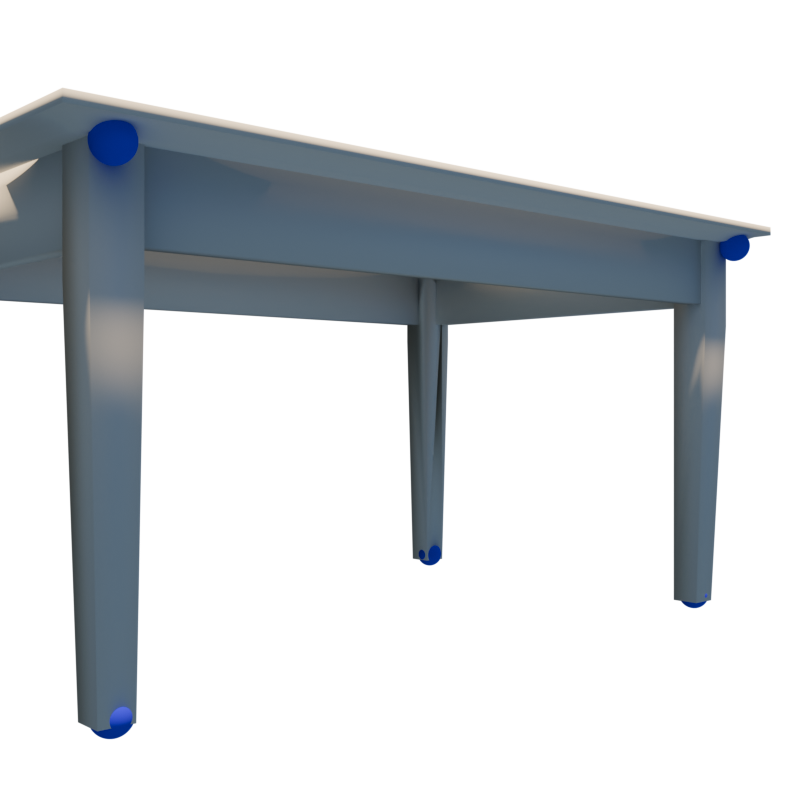}
		\caption{6 Views}
	\end{subfigure}
	\begin{subfigure}[t]{0.32\linewidth}
		\includegraphics[width=\textwidth,trim={0 10 0 15},clip]
		{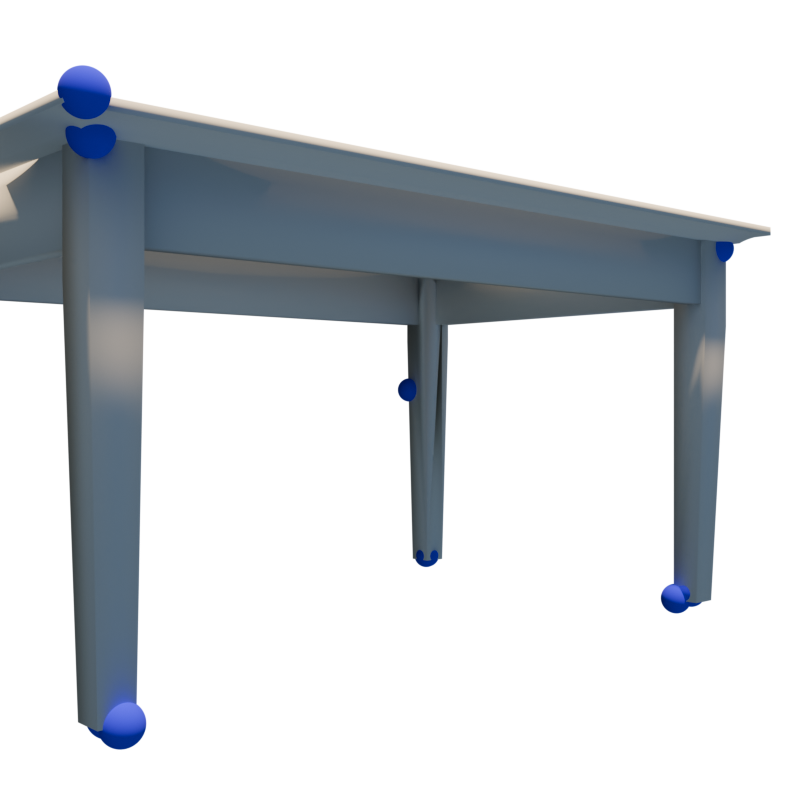}
		\caption{26 Views}
	\end{subfigure}
	\begin{subfigure}[t]{0.32\linewidth}
		\includegraphics[width=\textwidth,trim={0 10 0 15},clip]
		{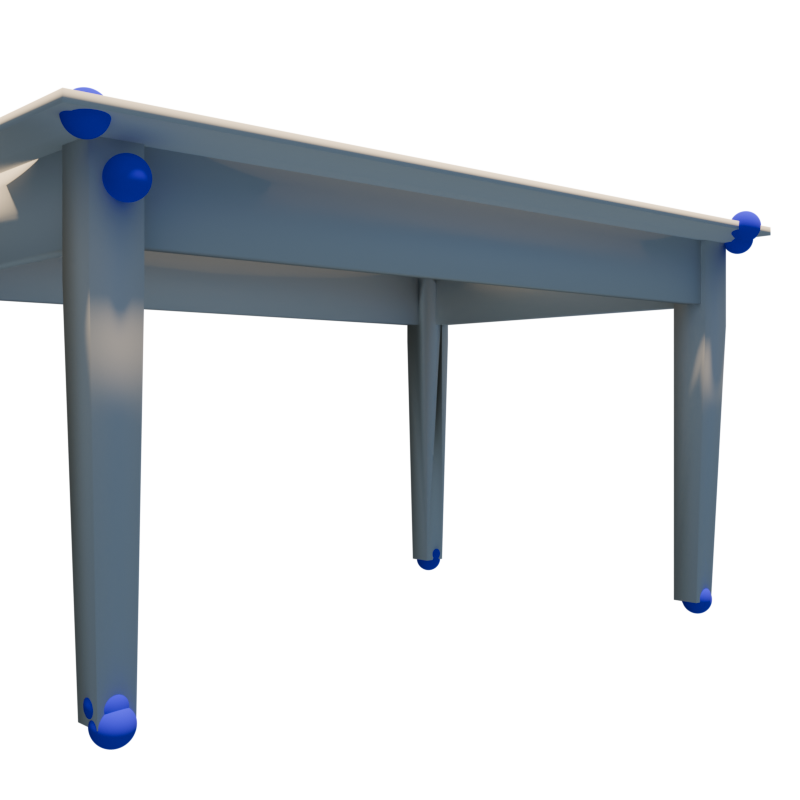}
		\caption{46 Views}
	\end{subfigure}
	\caption{The number of rendered views versus the detected keypoints after aggregation. This figure shows how varying the number of rendered views affects the total number of keypoints detected by ZeroKey. The prompt here is \textit{``corner of the table''}. As the number of views increases, ZeroKey will detect more keypoints that suit the description of the prompt. }
    \vspace{-1em}
	\label{fig:views}
\end{figure}

After obtaining the 2D keypoint detections from Molmo, we reconstruct the 3D coordinates of each keypoint in shape by back-projecting the 2D points into 3D space and aggregating the results from multiple views. Fig. \ref{fig:views} shows aggregated points from different numbers of views.

\mysection{Back-Projection of 2D Keypoints}
Given the camera projection matrices $\{\mathbf{C}_j\}_{j=1}^M$ corresponding to each image $\mathbf{V}_j$, we can back-project the 2D points $\mathbf{p}_{i,j}$ into 3D space.
Assuming a pinhole camera model, each camera projection matrix $\mathbf{C}_j$ maps 3D points $\mathbf{P} \in \mathbb{R}^3$ to 2D points $\mathbf{p} \in \mathbb{R}^2$:
\begin{equation}
\mathbf{p}_{i,j} \sim \mathbf{C}_j \mathbf{P}_{i,j}, \quad i = 1, \dots, N;\quad j = 1, \dots, M,
\end{equation}
where $\sim$ denotes equality up to scale.
To back-project $\mathbf{p}_{i,j}$ into 3D space, we compute the ray $\mathbf{r}_{i,j}(t)$ corresponding to $\mathbf{p}_{i,j}$:
\begin{equation}
\mathbf{r}_{i,j}(t) = \mathbf{C}_j^{-1}(\mathbf{p}_{i,j}, t),
\end{equation}
where $t$ represents the depth along the ray.
We intersect the ray with the 3D shape $S$ to find the 3D point $\mathbf{P}_{i,j}$:
\begin{equation}
\mathbf{P}_{i,j} = \text{Intersect}(\mathbf{r}_{i,j}, S).
\label{equ:intersect}
\end{equation}
%In our implementation, we predict keypoints based on the rendered views $\mathcal{V}$. 
During the rendering process, we cache the depth $t$ for each pixel and utilize that information for back-projection, eliminating the need for expensive intersection calculations.

\mysection{Aggregation of 3D Keypoints}
For each of the $N$ keypoint $k_i$, we collect the set of back-projected 3D points from all views:
$
\mathcal{P}_i = \{ \mathbf{P}_{i,j} ~|~ j = 1, \dots, M \}.
$
To obtain the final 3D keypoint location $\hat{\mathbf{P}}_i$, we aggregate the set $\mathcal{P}_i$. A straightforward method is to compute the mean of the points:
\begin{equation}
\hat{\mathbf{P}}_i = \frac{1}{|\mathcal{P}_i|} \sum_{j=1}^{M} \mathbf{P}_{i,j}.
\end{equation}
Naive summarization often produces unsatisfactory results. This occurs because back-projection can become unstable when the angle of the ray-mesh intersection is sharp. Molmo \cite{deitke2024molmo} can also generate multiple keypoints for one prompt, resulting in ambiguity and inconsistencies in the summarized data.

\mysection{Filtering and Refinement}
To handle sharp angular intersections and stabilize the mapping process, we back-project a  $h \times h$ patch $\mathbf{S}_{i,j}$ in the region centered at $\mathbf{P}_{i,j}$. We exclude all pixels that do not intersect with the mesh, then compute the mean of the back-projected points within the patch to obtain a stable back-projection of $\mathbf{P}_{i,j}$. Below is the revised back projection from Equation. \ref{equ:intersect}:
\begin{equation}
\mathbf{P}_{i,j} =\frac{1}{|\mathbf{S}_{i,j}|} \sum_{p,q\in \mathbf{S}_{i,j}} \text{Intersect}(\mathbf{r}_{p,q}, S).
\label{equ:intersect_patch}
\end{equation}
where $\mathbf{S}_{i,j}$ represents the set of all pixels within the patch whose rays intersect with the mesh $S$, and $|\mathbf{S}_{i,j}|$ denotes the number of such pixels.

\begin{figure*}[t]
	\centering
	\includegraphics[width=0.9\linewidth]{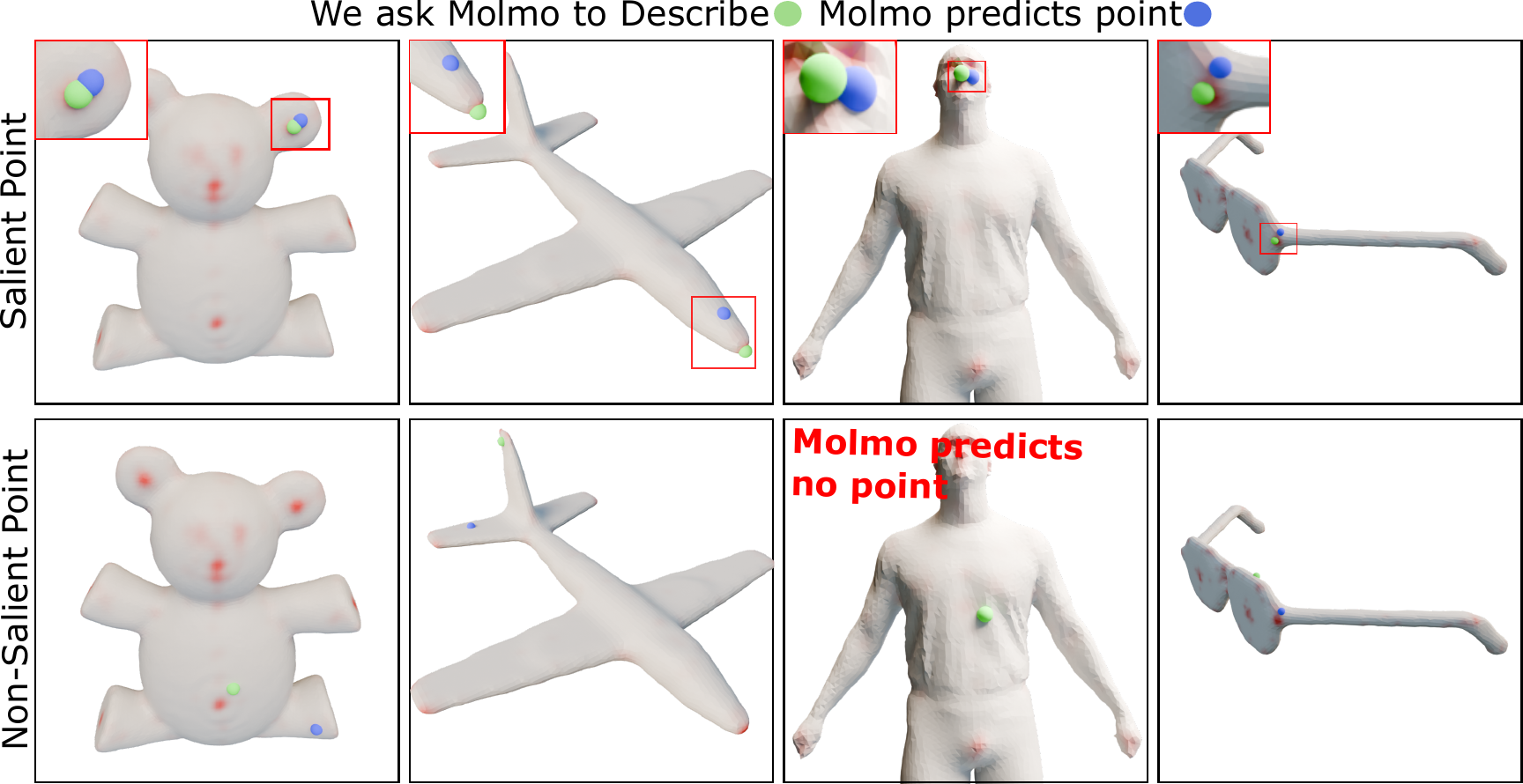}
	\caption{We ask Molmo to describe the green point, and using this as a prompt ZeroKey predicts the blue point. We show that salient points, given by the Schelling Points paper \cite{chen2012schelling}, are more easily describable and consistently retrievable than arbitrary points. Some arbitrary points even lead to ZeroKey being unable to find any suitable points.}
    \vspace{-1.5em}
	\label{fig:describe}
\end{figure*}

Patchify back projection effectively addresses the issue of sharp angular intersections. However, the predictions from Molmo can still yield multiple points for each prompt and may appear noisy across different views. One approach to improve this is to extract the confidence weighting from Molmo's feature map and apply it to the predictions. Nonetheless, normalizing the confidence scores across various views remains a challenge. In our method, we treat Molmo as a black box and employ hierarchical density-based spatial clustering on the predicted points obtained from multiple views.
Specifically, we apply HDBSCAN to the set of predicted points ${\mathcal{P}_{i}}$ obtained from multiple views to identify dense clusters that represent consistent predictions across views. HDBSCAN operates by defining the mutual reachability distance $d_{\text{m}}$ between two points $\mathbf{P}_{i,j}$ and $\mathbf{P}_{i,l}$ as:
\begin{equation}
d_{\text{m}} = \max\left(\text{core}_{k}(\mathbf{P}_{i,j}),\ \text{core}_{k}(\mathbf{P}_{i,l}),\ d(\mathbf{P}_{i,j}, \mathbf{P}_{i,l})\right),
\label{equ:mutual_reachability}
\end{equation}
where $\text{core}_{k}(\mathbf{P}_{i,j})$ is the core distance of point $\mathbf{P}_{i,j}$, defined as the distance to its $k$-th nearest neighbor, and $d(\mathbf{P}_{i,j}, \mathbf{P}_{i,l})$ is the Euclidean distance between points $\mathbf{P}_{i,j}$ and $\mathbf{P}_{i,l}$. By utilizing the mutual reachability distance and setting correct minPts $k$, we effectively filter out the noise and identify the most stable clusters in the predicted points.

\section{Applications}

Powerful VLMs, such as Molmo \cite{deitke2024molmo}, have strong 2D spatial reasoning abilities. We find that this reasoning generalizes to some degree to 3D. In this section, we explore some applications that explore the 3D knowledge encoded in Molmo.

\subsection{Schelling Points on 3D Surface}

Schelling points, as introduced by \citet{schelling1980strategy}, are specific focal points that people are likely to choose independently due to their prominence. In visual perception, these keypoints stand out due to distinct features. We hypothesize that the salient points people select often have clear, widely understood names in language. Therefore, an effective visual language model (VLM) should be able to better predict such salient points by leveraging its knowledge of both language and vision.

To illustrate this idea, we conducted experiments using a dataset \cite{chen2012schelling} that captures human-selected Schelling points. 
We utilized ZeroKey to generate a set of salient points for various images. The prompts guiding ZeroKey were provided by ChatGPT, simulating natural language descriptions that humans might use. This approach aimed to replicate the human process of identifying and naming keypoints based on visual and linguistic salience.
An illustration of this application is available in the supplementary material. We find in practice that the prompts given by ChatGPT align as expected with the most salient points reported by \cite{chen2012schelling}. 

\subsection{Point Describability and Consistency}

During our experiments, we observed that certain keypoints were more challenging to retrieve than others. This was raised when performing our quantitative analysis on the KeypointNet dataset \cite{you2020keypointnet}, which provides ground truth keypoints for various 3D models. Assigning a concise and descriptive text prompt to each ground truth point proved to be part of the challenge. We noticed a correlation: when a point was easy to name succinctly by a human observer, ZeroKey was able to retrieve it more effectively. Based on this intuition, we propose to study how describable different points are and how consistently ZeroKey can retrieve them.

To quantify how ``describable'' a point is, we leverage the distributions reported in the Schelling Points paper \cite{chen2012schelling}. Points with high-density values in the distributions reported by \cite{chen2012schelling} are considered more describable due to their inherent semantic significance. We refer to such points as salient points. The experiment is executed as shown in Alg.\ref{alg:ss_spp}.
\begin{algorithm}
\caption{Shape Selection and Saliency Prediction}
\label{alg:ss_spp}
\begin{algorithmic}[1]
    \Require Random shape $S$, and a point $p \in S$.
    \Comment Choose a point $p \in S$ with 1) the highest reported saliency value; or 2) without semantic meaning (saliency value is 0).
    \Ensure Absolute error shows that if $p$ is the saliency point
    \State Place a red marker at $p$ on $S$, render several views.
    \State Ask Molmo to describe the $p$ with a short name $l$, retaining the most repeated prompt.
    \State Render new views of $S$ without any marker.
    \Comment Query these views with prompt $l$.
    \State Infer a 3D point prediction $\hat{p}\leftarrow$ ZeroKey$(S,l)$.  
    \State Store the absolute error $e\leftarrow|\hat{p}-p|$.
\end{algorithmic}
\end{algorithm}

\begin{figure}
    \centering
    \includegraphics[width=\linewidth,trim={0 5 0 5},clip]{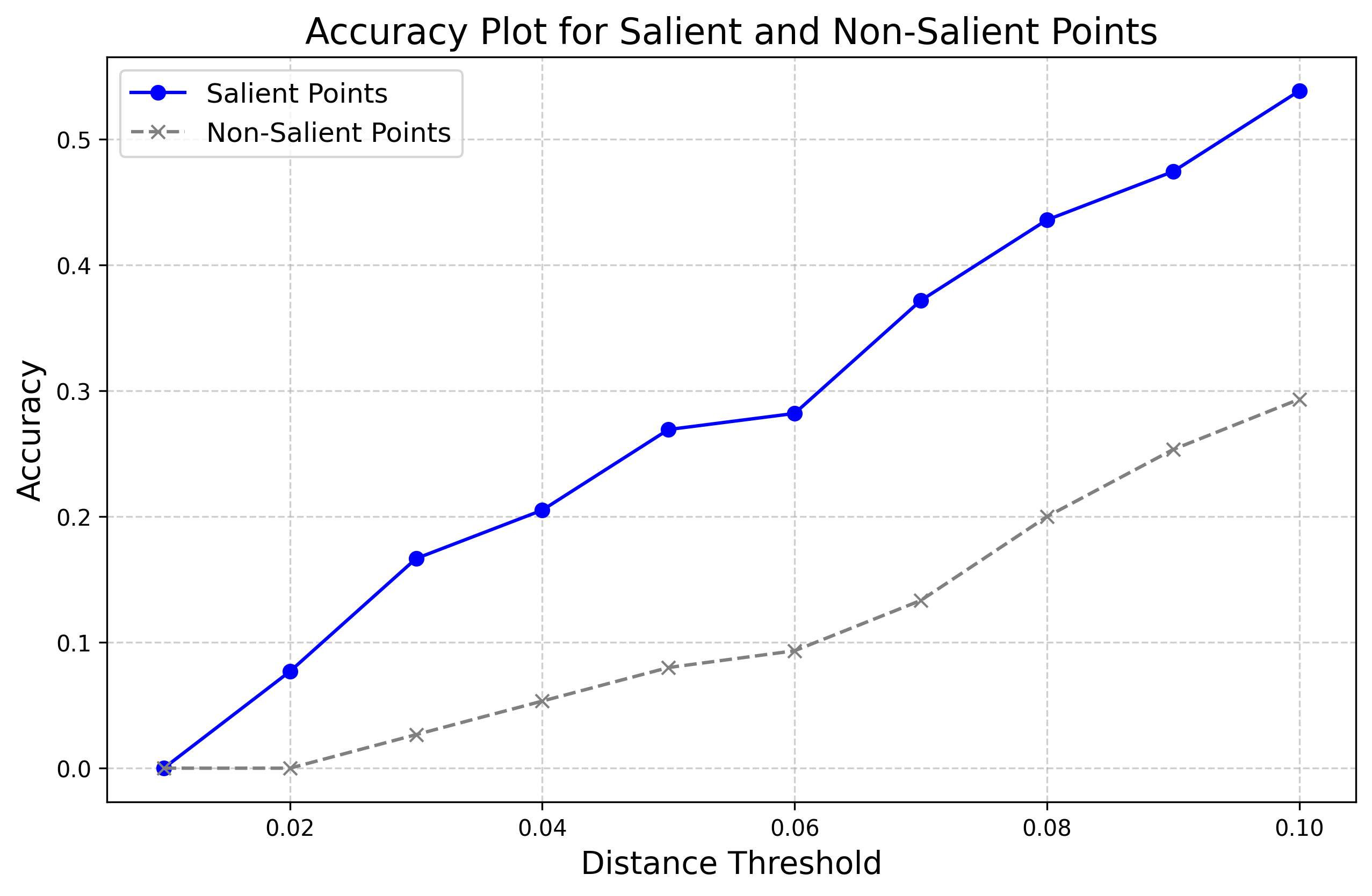}
    \caption{We show through a quantitative study that ``salient" points are retrieved with higher accuracy than ``non-salient" ones, regardless of the distance threshold used.}
    \vspace{-1.5em}
    \label{fig:schelling_points_comparison}
\end{figure}

The qualitative results of our experiments are showcased in Fig.~\ref{fig:describe}. For this experiment we ask Molmo to describe the green dot, using a simple prompt. We then give the prompt generated by Molmo to Zerokey to predict a 3D location of the original point. We can see that for salient points, ZeroKey is able to leverage the prompt given by Molmo to relocate the original queried point with good accuracy. However, for other ``random" non-salient points this tasks is much more complicated. ZeroKey points to similar parts of the object, such as the airplane wing; or seemingly unrelated parts, like for the teddy bear. Moreover, we observed a fail state where for some non-salient points Molmo was unable to give a good prompt, resulting in ZeroKey affirming that no points matching the prompt were present in the scene.
Quantitative results can be found in Fig.~\ref{fig:schelling_points_comparison}. In this figure, we see that the accuracy of retrieving a point from a Molmo description using Molmo is much higher for a semantic meaningful point than an arbitrary point. 

\begin{table*}[t]
\centering
\vspace{-1.5em}
\begin{tabular}{|c|c|c|c|c|c|c|c|c|c|c|c|}
\hline
Method & \multicolumn{11}{c|}{IoU (in \%)} \\ \hline
Distance Thresholds & 0.001 & 0.01 & 0.02 & 0.03 & 0.04 & 0.05 & 0.06 & 0.07 & 0.08 & 0.09 & 0.10 \\ \hline
RedCircle \cite{shtedritski2023does} & 0.21 & 0.34 & 0.64 & 1.16 & 1.90 & 3.05 & 4.81 & 7.55 & 11.06 & 14.92 & 18.50 \\
GPT-4o & 0.18 & 0.48 & 1.20 & 2.61 & 4.19 & 6.04 & 8.48 & 10.89 & 14.03 & 17.03 & 20.73 \\
CLIP-DINOiser \cite{wysoczanska2024clip} & 0.73 & 1.41 & 3.00 & 4.94 & 7.31 & 9.80 & 12.66 & 15.52 & 18.52 & 21.76 & 25.56 \\ \hline
\textbf{ZeroKey (Ours)} & \textbf{3.69} & \textbf{5.84} & \textbf{10.91} & \textbf{17.58} & \textbf{23.72} & \textbf{29.81} & \textbf{35.52} & \textbf{40.82} & \textbf{45.72} & \textbf{50.24} & \textbf{54.64} \\ \hline
\end{tabular}
\caption{Comparison of IoU between the predicted and ground-truth keypoints from KeypointNet using different methods across various geodesic distance thresholds.}
\vspace{-1.5em}
\label{tab:comparison}
\end{table*}

\section{Experiments}

\subsection{Setup and Dataset}
We evaluate our method using the KeypointNet dataset.
Our evaluation strategy follows the approach of \citet{you2022ukpgan}, which computes the Intersection over Union (IoU) between predicted keypoints and ground-truth keypoints from the KeypointNet dataset, using varying distance thresholds. A match is counted if the geodesic distance between a ground-truth keypoint and a predicted keypoint is less than the specified threshold. We focus on the same three classes from the dataset during our evaluation: airplane, chair, and table.
The labels in KeypointNet only provide integer IDs for each keypoint. To detect keypoints from open vocabulary textual descriptions and evaluate them using KeypointNet, we manually annotated the labels of keypoint classes in the dataset. We then used this annotated text as a textual prompt for zero-shot 3D keypoint detection. Please refer to our supplementary for the exact prompt we used in each category.

\begin{figure}[t]
    \centering
    \includegraphics[width=\linewidth]{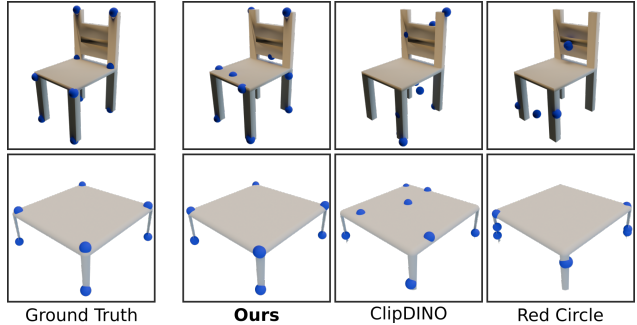}
    \caption{We compare against baselines CLIP-DINOiser and RedCircle. While both baselines identify some prominent regions, they fall in accurately localizing keypoints according to the text prompt. In contrast, our method effectively detects and precisely locates keypoints.}
    \label{fig:enter-label}
    \vspace{-1em}
\end{figure}

\subsection{Zero-Shot 3D Keypoint Detection Baseline}
% Description of experiments
\mysection{RedCircle} The Red Circle method \cite{shtedritski2023does} is one of the simplest and most straightforward ways to highlight points in the image based on a text query by randomly sampling points and highlighting them with a red circle and picking the one point with the highest CLIP similarity to the text query. We lift the prediction of this method to 3D using the same lifting procedure used in our method to compare 3D Zero-shot keypoint detection.  

\mysection{CLIP-DINOiser} The CLIP-DINOiser \cite{wysoczanska2024clip} combines the strengths of CLIP and DINO to perform zero-shot object localization and segmentation based on text query in images. To adapt the CLIP-DINOiser for zero-shot 3D keypoint detection, we first apply the method to each view of the 3D scene to obtain 2D keypoint predictions corresponding to the text prompt. For each image, the CLIP-DINOiser produces a heatmap indicating the regions of interest related to the query. We select the point with the highest activation in the heatmap as the 2D keypoint prediction in that view. We then lift these 2D keypoints to 3D using the same back-projection technique described in our method. 

\mysection{GPT-4o}
In this Baseline, we replace our model with GPT-4o to observe variations in performance. Notably, we highlight that MLLMs trained solely on image-level tasks fail to provide useful signals for keypoint detection. In the GPT-4o version, we replace Molmo model in our method with the latest GPT-4o (2024-08-06) and evaluate its performance on point-level tasks. This comparison highlights the necessity of specialized training or architectural features that support fine-grained reasoning in visual contexts.

\subsection{Quantitative and Qualitative Analysis}
In our evaluation of KeypointNet, our Zero-Shot method significantly outperforms the VLM-based baselines across all distance thresholds. It achieves IoU levels that are comparable to those of reference-based Few-Shot methods and supervised methods specifically tailored for this dataset. Qualitative results are shown in Table \ref{tab:comparison}. In Fig.~\ref{fig:all_shapes} we show side by side comparisons between some ground truth keypoints and our Zero-Shot prediction.

\subsection{Ablation Studies}

\begin{figure}
    \centering
    \includegraphics[width=\linewidth,trim={0 10 0 0},clip]{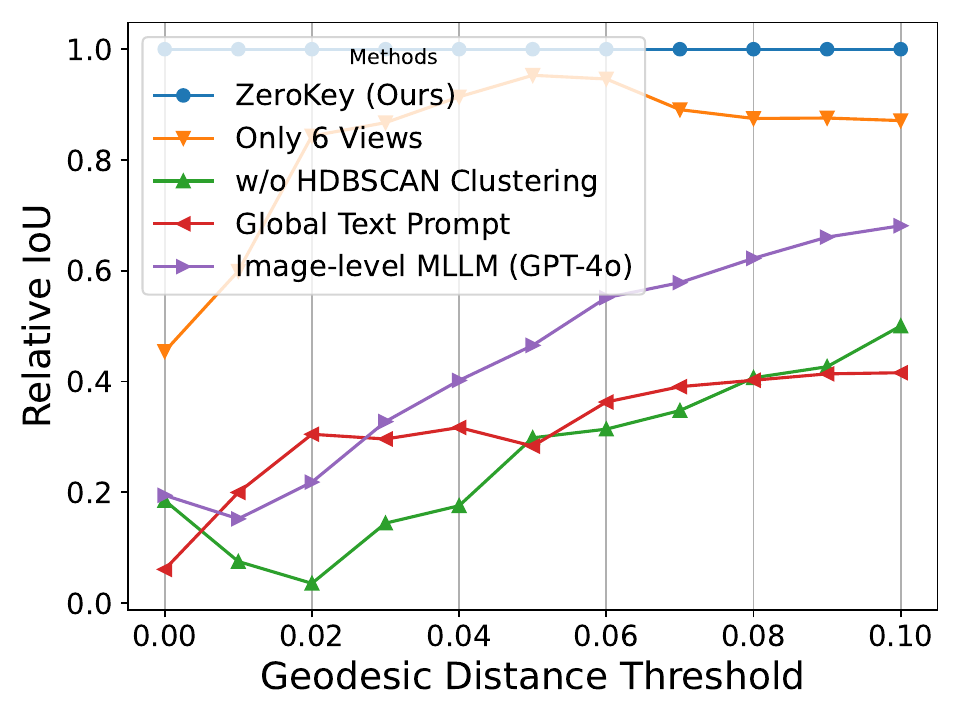}
    \caption{Comparison of the performance across different configurations: (blue) our original method; (green) results with a Global Text prompt; (red) results using GPT-4 as the MLLM backbone; and (yellow) results without HDBSCAN clustering. This figure shows the impact of each modification on the overall performance.}
    \vspace{-1.5em}
    \label{fig:ablation}
\end{figure}

In this ablation study, we modify different components of our method to gain deeper insights into the point-level reasoning abilities of MLLMs. Specifically, we compare the effects of various text prompts, experiment with different MLLM backbones, and evaluate the impact of aggregating information from different numbers of views.

\mysection{Global Text prompt} We evaluate the effectiveness of a point-specific text prompt. The original method utilizes prompts tailored to specific points in the image, providing precise guidance to the model. In the ablated version, we replace these point-specific text prompts with a global prompt: ``Point to all salient points in this image.'' We then compare the detected set of keypoints with the ground truth to assess how this change affects the model's ability to localize specific points. This experiment helps us understand the importance of prompt specificity in guiding the MLLM's attention to particular regions of the image.

\begin{figure}
    \centering
    %mesh_id: fbd48960edc73ef0490ad276cd2af3a4
    \includegraphics[width=\linewidth,trim={20 90 20 20},clip]{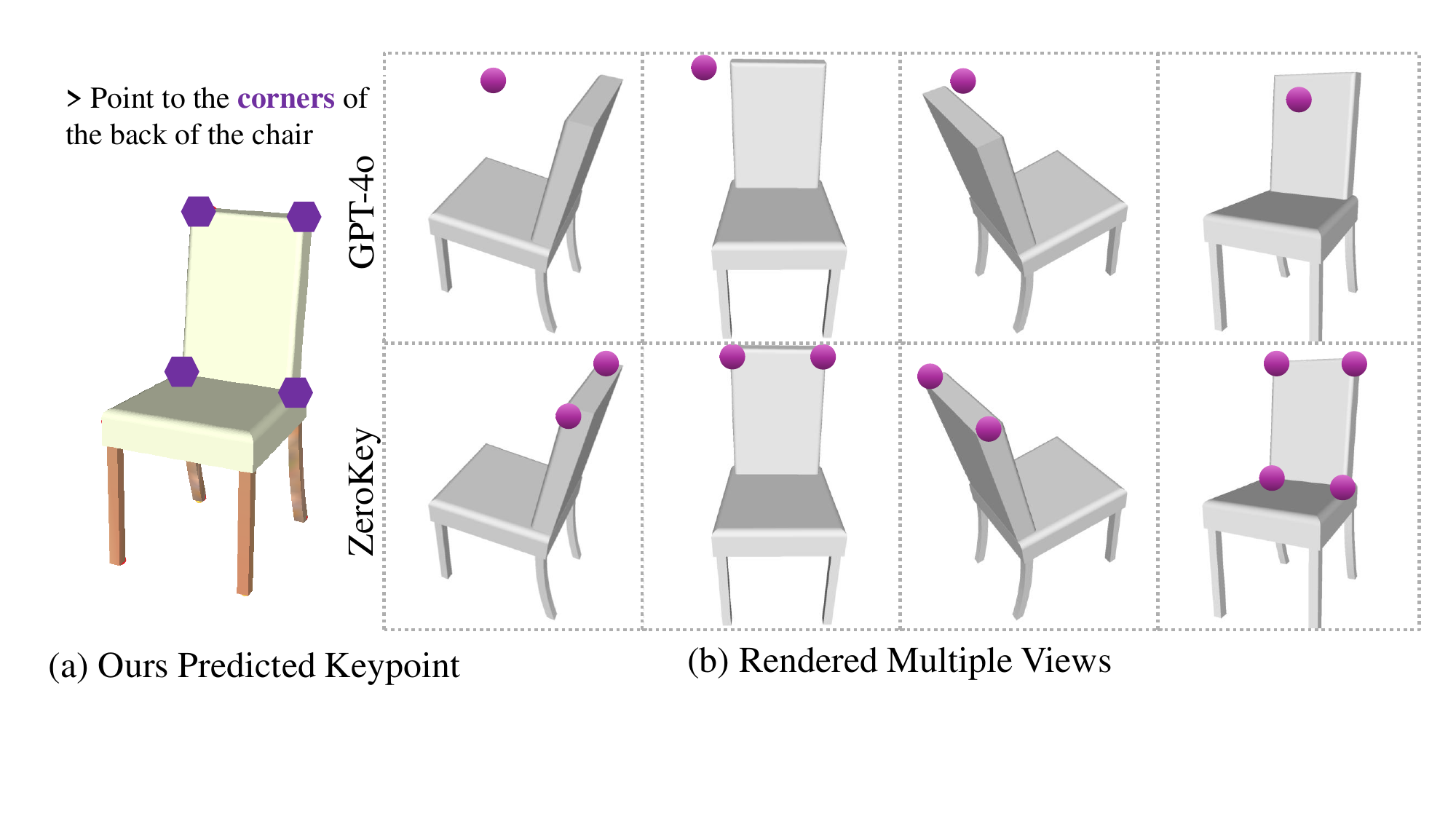}
    \caption{Renderings from the ablation study where the MLLM backbone is replaced with GPT-4o. As an MLLM trained solely on image-level tasks, it struggles with point-level reasoning, leading to less accurate and consistent keypoints compared to Molmo. }
    \vspace{-1.5em}
    \label{fig:molmo_vs_gpt4o}
\end{figure}

\mysection{Aggregation ablation} 
We explore aggregating information from multiple views to enhance the model's understanding of the scene. Our original method aggregates keypoint information from 26 viewpoints to improve accuracy. In the 6-views variant, we reduce the number of viewpoints to 6. In another variant, we compare the performance of direct average against HDBSCAN clustering for aggregating detections. By evaluating these different aggregation strategies, we aim to determine how the number of views and the method of combining information influence the model's ability to accurately identify keypoints.

The ablation results shown in Fig. \ref{fig:ablation} illustrate the relative performance compared with ZeroKey. The ability to achieve 80\% performance with only 6 views indicates that our method efficiently aggregates points. The poor performance of direct averaging suggests the necessity of HDBSCAN clustering. Fig. \ref{fig:molmo_vs_gpt4o} demonstrates that GPT-4o fails to precisely locate the keypoint. This shows the need for an MLLM backbone trained with point-level tasks for precise keypoint detection.
\section{Conclusion and Future Work}
In this paper, we presented a novel zero-shot approach for 3D keypoint detection by harnessing the capabilities of recent MLLMs trained with enhanced localized reasoning capabilities. 
Unlike traditional methods that rely heavily on annotated 3D datasets and extensive supervised training, our method leverages the pixel-level annotations inherent in MLLMs to both extract and name salient keypoints without any ground truth labels or supervision. 
Our experimental evaluations demonstrate that our approach achieves remarkable performance. 
We believe that this work opens the door to entirely novel applications by providing a bridge between 3D data and MLLMs with \textit{localized} reasoning. 
This can help solve downstream tasks such as shape manipulation, deformation, and processing, which typically require significant domain knowledge. Furthermore, this link can also help to enhance the power of MLLMs by providing feedback thanks to multiview consistency from 3D data, thus potentially helping to improve the robustness of their pixel-level understanding.

\mysection{Acknowledgments}
Parts of this work were supported by the ERC Consolidator Grant 101087347 (VEGA) and the ANR AI Chair AIGRETTE. This work was also supported by the KAUST Center of Excellence for Generative AI, under award number 5940, and the KAUST Ibn Rushd Postdoc Fellowship program. 
{
    \small
    \bibliographystyle{ieeenat_fullname}
    \bibliography{main}
}

\clearpage \clearpage
\appendix 
\clearpage
\maketitlesupplementary

\section{3D Multimodal Language Models}
\label{sec:3dllm}
Recent general-purpose 3D language models (LLMs) are usually trained with specific objectives that align 3D data representations with vision-language models, allowing for global tasks like classification. However, these models are based on standard vision-language frameworks that have limited localization capabilities. Consequently, they are unable to accurately perform 3D keypoint localization based on language prompts.
\begin{figure}
    \centering
    %mesh_id: fbd48960edc73ef0490ad276cd2af3a4
    \includegraphics[width=\linewidth,trim={0 40 500 0},clip]{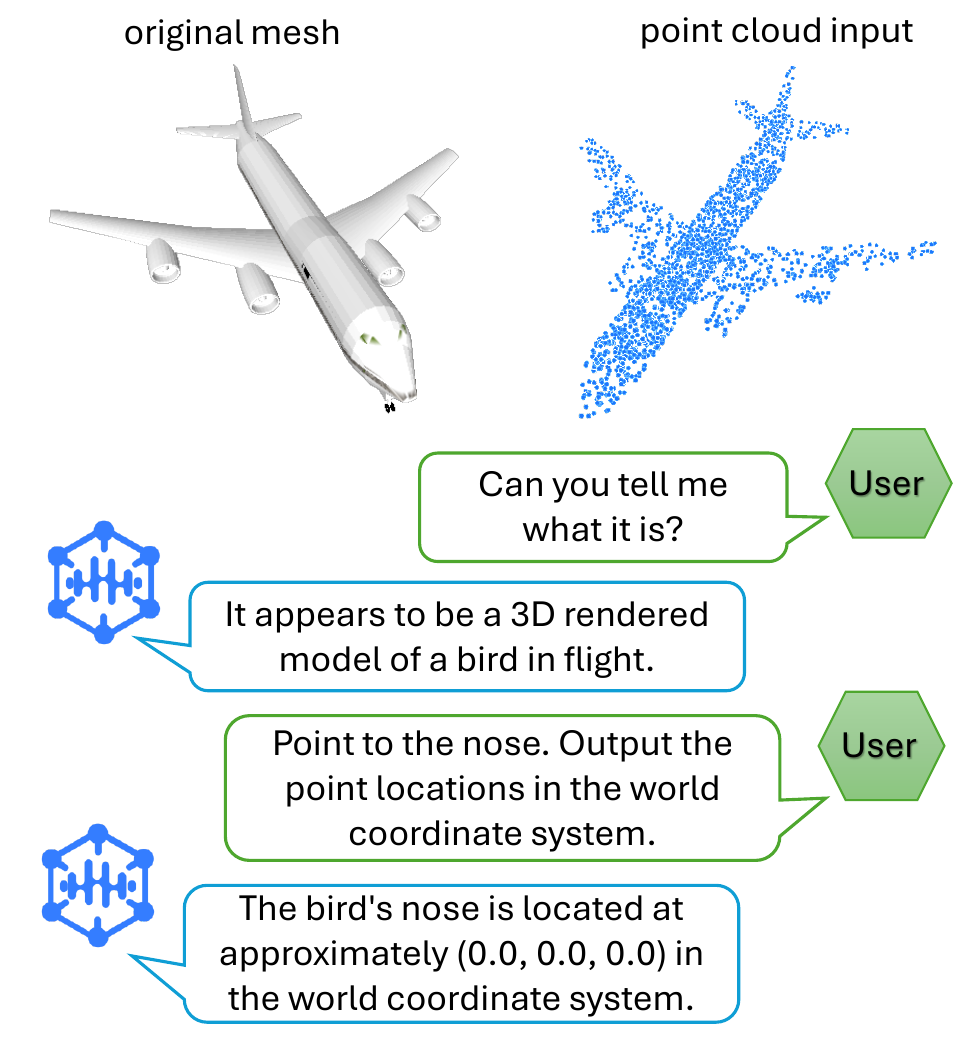}
    \caption{Qualitative comparison of keypoint localization using ShapeLLM on out-of-distribution 3D airplane shape. ShapeLLM struggles to accurately identify the shape and point to keypoints on this unseen pointcloud data.}
    \vspace{-1.5em}
    \label{fig:shapellm}
\end{figure}
We evaluated the KeypointNet model against the state-of-the-art ShapeLLM \cite{qi2025shapellm}. The qualitative results are shown in Fig.~\ref{fig:shapellm}. We observed that 3D LLMs are often sensitive to the alignment and rotation of the input point cloud and do not generalize well to new datasets. Even with perfectly aligned point clouds, we found that ShapeLLM still struggled to accurately understand out-of-distribution shapes and localize keypoints when the shapes were not seen during its training phase. These limitations highlight the challenges faced by current 3D language models in precise localization and generalization to unseen data in zero-shot settings.

\section{Schelling Points}

\begin{figure}
    \centering
    \includegraphics[width=\linewidth]{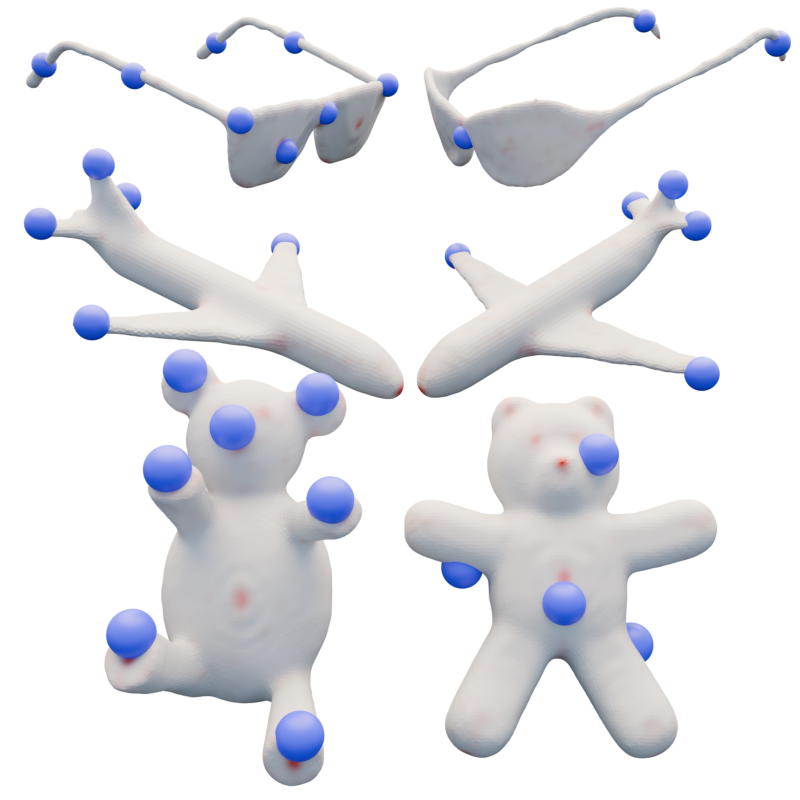}
    \caption{Superposition of the keypoints predicted by ZeroKey (blue) and the Schelling point distribution \cite{chen2012schelling} (red).}
    \label{fig:schelling_points}
\end{figure}

%We make the hypothesis that the salient points people select often have clear names in language. Therefore, we believe that an effective visual language model (VLM) should be able to better predict such salient points by leveraging its knowledge of both language and vision.
We make the hypothesis that the salient points selected by individuals in visual scenes often correspond to objects or features with clear, identifiable names in natural language. This leads us to believe that an effective Visual Language Model (VLM) should be capable of predicting such salient points by leveraging its combined understanding of language and vision.

%To illustrate this idea, we conducted experiments using a dataset \cite{chen2012schelling} that captures human-selected Schelling points. 
To test this hypothesis, we conducted experiments using the dataset from \citet{chen2012schelling}, which captures human-selected Schelling points in various visual scenarios. In the context of coordination games, Schelling points refer to prominent locations or choices that people commonly select without prior communication, relying on mutual expectations and shared cultural references.

\begin{table*}[t]
\centering
\vspace{-1.5em}
\begin{tabular}{|c|c|c|c|c|c|c|c|c|c|c|c|}
\hline
Method & \multicolumn{11}{c|}{IoU (in \%)} \\ \hline
Distance Thresholds & 0.001 & 0.01 & 0.02 & 0.03 & 0.04 & 0.05 & 0.06 & 0.07 & 0.08 & 0.09 & 0.10 \\ \hline
HARRIS-3D & 0.15 & 0.76 & 2.19 & 3.96 & 6.16 & 8.88 & 11.91 & 15.13 & 18.63 & 22.10 & 25.69 \\
SIFT-3D & 0.29 & 1.05 & 2.62 & 4.38 & 6.65 & 9.42 & 12.38 & 15.65 & 19.39 & 22.71 & 26.15 \\
ISS & 0.49 & 1.10 & 2.69 & 4.67 & 7.27 & 10.14 & 13.29 & 16.56 & 20.02 & 23.58 & 27.20 \\
USIP & 0.83 & 1.70 & 3.25 & 5.24 & 8.00 & 11.76 & 15.98 & 20.63 & 25.56 & 30.20 & 33.67 \\
D3FEAT & 2.36 & 3.86 & 7.27 & 10.37 & 13.37 & 17.12 & 21.41 & 26.12 & 30.39 & 34.82 & 38.41 \\
UKPGAN & 3.95 & 6.54 & 12.27 & 17.86 & 22.42 & 26.55 & 30.28 & 34.32 & 38.46 & 42.65 & 46.49 \\
FPS & 6.34 & 10.55 & 17.68 & 23.18 & 26.91 & 30.27 & 33.57 & 36.85 & 40.47 & 44.42 & 48.38 \\
FSKD & 7.00 & 7.94 & 11.17 & 16.74 & 23.99 & 31.14 & 38.14 & 43.97 & 49.30 & 53.62 & 57.03 \\
B2-3D (62 views) & 12.41 & 20.29 & 35.55 & 46.25 & 52.74 & 57.72 & 61.61 & 64.19 & 66.56 & 68.55 & 70.57 \\
B2-3D (26 views) & 8.60 & 14.39 & 25.10 & 33.31 & 39.29 & 44.11 & 47.85 & 50.83 & 53.17 & 55.11 & 56.90 \\
\hline
StablePoints \cite{hedlin2024unsupervised} & 3.66 & 5.80 & 9.94 & 13.32 & 16.58 & 19.91 & 23.48 & 27.26 & 31.28 & 34.89 & 38.22 \\ \hline
RedCircle \cite{shtedritski2023does} & 0.21 & 0.34 & 0.64 & 1.16 & 1.90 & 3.05 & 4.81 & 7.55 & 11.06 & 14.92 & 18.50 \\
GPT-4o & 0.18 & 0.48 & 1.20 & 2.61 & 4.19 & 6.04 & 8.48 & 10.89 & 14.03 & 17.03 & 20.73 \\
CLIP-DINOiser \cite{wysoczanska2024clip} & 0.73 & 1.41 & 3.00 & 4.94 & 7.31 & 9.80 & 12.66 & 15.52 & 18.52 & 21.76 & 25.56 \\ \hline
\textbf{ZeroKey (Ours)} & \textbf{3.69} & \textbf{5.84} & \textbf{10.91} & \textbf{17.58} & \textbf{23.72} & \textbf{29.81} & \textbf{35.52} & \textbf{40.82} & \textbf{45.72} & \textbf{50.24} & \textbf{54.64} \\ \hline
\end{tabular}
\caption{Comparison of IoU between the predicted and ground-truth keypoints from KeypointNet using different methods across various geodesic distance thresholds. The IoU of our method is highlighted in bold.}
\vspace{-1.5em}
\label{tab:comparison_suppl}
\end{table*}

In Fig.~\ref{fig:schelling_points}, we showcase a visualization that superposes the distributions reported by the Schelling Points paper \cite{chen2012schelling} to a set of salient points generated by ZeroKey, where the candidate keypoints were provided by ChatGPT using the following prompt:
\begin{center}
    \textit{``List possible salient keypoints (in text).''}
\end{center}
Besides text prompt, we also input renderings of the 3D scene to ChatGPT to obtain prompts simulates natural language descriptions that humans are likely to use. This process can simulate the way humans might describe these scenes in natural language, capturing the linguistic cues that people are likely to use when identifying salient features.

We see in this image that some of the salient points \cite{chen2012schelling} do correspond to the predictions made by ZeroKey. Our findings reveal that several of the salient points identified by humans correspond to the predictions made by ZeroKey. These results support our hypothesis that salient points in visual scenes are closely tied to language and that VLMs that harness this connection can more effectively predict human-selected points of interest.

\section{More Quantitative Analysis}

In this supplementary document, we present additional quantitative evaluations. We compared our Zero-Shot method with both supervised and few-shot methods trained on KeypointNet. Our results demonstrate that the Zero-Shot method can even outperform some of the supervised methods. The IoU levels achieved by our method are comparable to those of the reference-based Few-Shot methods and the supervised methods specifically designed for this dataset. Qualitative results are provided in Table \ref{tab:comparison_suppl}. Our method is evaluated using 26 rendered views, and we have also included the results from the same number of views of the few-shot method B2-3D as a reference.

In addition to the supervised and few-shot methods, we also compared our approach with an unsupervised baseline called StablePoint \cite{hedlin2024unsupervised}. This method learns keypoints by optimizing text embeddings from latent diffusion models. The main idea is to identify text embeddings that guide the generative model to consistently focus on compact regions within images, which are then used as keypoints. Unlike our approach, this baseline does not retrieve keypoint according to a text prompt and operates fully unsupervised. However, it does necessitate specifying the exact number of keypoints we wish to detect. In our experiments, we set the number of keypoints to match the number of ground-truth points in the KeypointNet dataset.

\section{Keypoint Descriptions for Categories}

\begin{table*}
\centering
\begin{tabular}{|l|l|l|l|}
\hline
Object & Index & Description & Short Description \\
\hline
\multirow{11}{*}{Table} & 0 & corner of the table top & different corners  \\
 & 1 & corner of the table top & different corners  \\
 & 2 & corner of the table top & different corners  \\
 & 3 & corner of the table top & different corners  \\
 & 4 & end of the table leg & end of the legs  \\
 & 5 & end of the table leg & end of the legs  \\
 & 6 & end of the table leg & end of the legs  \\
 & 7 & end of the table leg & end of the legs  \\
 & 8 & end of the stretcher between the table legs & - \\
 & 9 & end of the stretcher between the table legs & - \\
 & 10 & end of the stretcher between the table legs & - \\
 & 11 & end of the stretcher between the table legs & - \\
\hline
\multirow{10}{*}{Mug} & 0 & mug rim & - \\
 & 1 & mug rim & - \\
 & 2 & mug rim & - \\
 & 3 & mug rim & - \\
 & 4 & mug base rim & - \\
 & 5 & mug base rim & - \\
 & 6 & mug base rim & - \\
 & 7 & mug base rim & - \\
 & 8 & upper attachment point for the mug handle & top of the handle  \\
 & 9 & center of the mug handle & handle  \\
 & 10 & lower attachment point for the mug handle & bottom of the handle  \\
\hline
\multirow{17}{*}{Bottle} & 0 & lower rim of the bottle & - \\
 & 1 & lower rim of the bottle & - \\
 & 2 & lower rim of the bottle & - \\
 & 3 & lower rim of the bottle & - \\
 & 4 & edge between the bottle body and the bottle neck & - \\
 & 5 & edge between the bottle body and the bottle neck & - \\
 & 6 & edge between the bottle body and the bottle neck & - \\
 & 7 & edge between the bottle body and the bottle neck & - \\
 & 8 & bottle cap center & - \\
 & 9 & edge of the bottle cap & - \\
 & 10 & edge of the bottle cap & - \\
 & 11 & edge of the bottle cap & - \\
 & 12 & edge of the bottle cap & - \\
 & 13 & lower edge of the bottle neck & - \\
 & 14 & lower edge of the bottle neck & - \\
 & 15 & lower edge of the bottle neck & - \\
 & 16 & lower edge of the bottle neck & - \\
 & 17 & lower edge of the bottle neck & - \\
\hline
\end{tabular}
\end{table*}

\begin{table*}
\centering
\begin{tabular}{|l|l|l|l|}
\hline
Object & Index & Description & Short Description \\
\hline
\multirow{23}{*}{Vessel} & 0 & bow of a boat & front  \\
 & 1 & lower part of the end of the bow of a boat (starboard) & bottom front  \\
 & 2 & upper part of the end of the bow of a boat (starboard) & - \\
 & 3 & upper part of the end of the bow of a boat (port) & - \\
 & 4 & lower part of the end of the bow of a boat (port) & bottom front  \\
 & 5 & lower corner of the boat stern (starboard & back  \\
 & 6 & upper corner of the boat stern (starboard) & back left corner  \\
 & 7 & upper corner of the boat stern (port) & back left corner  \\
 & 8 & lower corner of the boat stern (port) & back  \\
 & 9 & mast step of a sailboat & - \\
 & 10 & mast top of a sailboat & - \\
 & 11 & lower front corner of the deckhouse (starboard) & right front of the cabin  \\
 & 12 & lower front corner of deckhouse (port) & left front of the cabin  \\
 & 13 & lower aft corner of the deckhouse (starboard) & - \\
 & 14 & lower aft corner of the deckhouse (port) & - \\
 & 15 & upper front corner of the deckhouse (starboard) & - \\
 & 16 & upper front corner of the deckhouse (port) & - \\
 & 17 & upper aft corner of the deckhouse (starboard) & - \\
 & 18 & upper aft corner of the deckhouse (port) & - \\
 & 19 & foremost elevation point on the deck of the ship & - \\
 & 20 & lower corner of the windshield of the ship (starboard) & - \\
 & 21 & lower corner of the windshield of the ship (port) & - \\
 & 22 & upper corner of the windshield of the ship (starboard) & - \\
 & 23 & upper corner of the windshield of the ship (port) & - \\
\hline
\multirow{20}{*}{Chair} & 0 & upper corner of the back of the chair & top corners of the back rest  \\
 & 1 & upper corner of the back of the chair & top corners of the back rest  \\
 & 2 & lower corner of the back of the chair & different corners of the seat  \\
 & 3 & lower corner of the back of the chair & different corners of the seat  \\
 & 4 & front corner of the seat of the chair & different corners of the seat  \\
 & 5 & front corner of the seat of the chair & different corners of the seat  \\
 & 6 & front corner of the armrest of the chair & end of both armrests  \\
 & 7 & back corner of the armrest of the chair & end of both armrests  \\
 & 8 & front corner of the armrest of the chair & end of both armrests  \\
 & 9 & back corner of the armrest of the chair & end of both armrests  \\
 & 10 & lower end of the front leg of the chair & tip of all legs  \\
 & 11 & lower end of the back leg of the chair & tip of all legs  \\
 & 12 & lower end of the front leg of the chair & tip of all legs  \\
 & 13 & lower end of the back leg of the chair & tip of all legs  \\
 & 14 & lower end of the leg of the chair & tip of all legs  \\
 & 15 & top of the chair cylinder & cylinder  \\
 & 16 & bottom of the chair cylinder & cylinder  \\
 & 17 & lower end of the front leg of the chair & - \\
 & 18 & lower end of the back leg of the chair & - \\
 & 19 & lower end of the back leg of the chair & - \\
 & 20 & lower end of the front leg of the chair & - \\
\hline
\end{tabular}
\end{table*}

\begin{table*}
\centering
\begin{tabular}{|l|l|l|l|}
\hline
Object & Index & Description & Short Description \\
\hline
\multirow{19}{*}{Bed} & 0 & lower end of the front leg of the bed & - \\
 & 1 & lower end of the headboard leg of the bed & - \\
 & 2 & lower end of the headboard leg of the bed & - \\
 & 3 & lower end of the front leg of the bed & - \\
 & 4 & foot corner of the mattress & - \\
 & 5 & head corner of the mattress & - \\
 & 6 & head corner of the mattress & - \\
 & 7 & foot corner of the mattress & - \\
 & 8 & foot corner of the upper mattress of the bunk bed & - \\
 & 9 & head corner of the upper mattress of the bunk bed & - \\
 & 10 & head corner of the upper mattress of the bunk bed & - \\
 & 11 & foot corner of the upper mattress of the bunk bed & - \\
 & 12 & upper corner of the bed frame of the bunk bed & - \\
 & 13 & upper corner of the bed frame of the bunk bed & - \\
 & 14 & upper corner of the bed frame of the bunk bed & - \\
 & 15 & upper corner of the bed frame of the bunk bed & - \\
 & 16 & upper corner of the footboard of the bed & - \\
 & 17 & upper corner of the headboard of the bed & - \\
 & 18 & upper corner of the headboard of the bed & - \\
 & 19 & upper corner of the footboard of the bed & - \\
\hline
\multirow{9}{*}{Skateboard} & 0 & corner of the tail of the skateboard & - \\
 & 1 & corner of the tail of the skateboard & - \\
 & 2 & corner of the nose of the skateboard & - \\
 & 3 & corner of the nose of the skateboard & - \\
 & 4 & wheel of the skateboard & wheel  \\
 & 5 & wheel of the skateboard & wheel  \\
 & 6 & wheel of the skateboard & wheel  \\
 & 7 & wheel of the skateboard & wheel  \\
 & 8 & nose of the skateboard & end  \\
 & 9 & tail of the skateboard & end  \\
\hline
\multirow{5}{*}{Knife} & 0 & tip of the knife & tip of the blade  \\
 & 1 & butt of the knife & bottom  \\
 & 2 & shoulder of the knife & - \\
 & 3 & heel of the knife & - \\
 & 4 & end of the crossguard of the knife & - \\
 & 5 & end of the crossguard of the knife & - \\
\hline
\multirow{5}{*}{Cap} & 0 & top button of the cap & top  \\
 & 1 & front of the visor of the cap & front of the visor  \\
 & 2 & corner of the visor of the cap & - \\
 & 3 & corner of the visor of the cap & - \\
 & 4 & center of the back of the cap & middle of the strap  \\
 & 5 & center of the visor attachment of the cap & - \\
\hline
\multirow{5}{*}{Laptop}   & 0 & corner of the laptop body & bottom left corner  \\
 & 1 & hinge of the laptop & bottom left corner of the screen  \\
 & 2 & hint of the laptop & bottom right corner of the screen  \\
 & 3 & corner of the laptop body & bottom right corner  \\
 & 4 & corner of the laptop screen & top left corner of the screen  \\
 & 5 & corner of the laptop screen & top right corner of the screen  \\
\hline
\end{tabular}
\end{table*}

\begin{table*}
\centering
\begin{tabular}{|l|l|p{6cm}|p{6cm}|}
\hline
Object & Index & Description & Short Description \\
\hline
\multirow{17}{*}{Motorcycle} & 0 & headlight of the motorcycle & headlight  \\
 & 1 & end of the handlebar of the motorcycle & right part of the steering wheel  \\
 & 2 & end of the handlebar of the motorcycle & left part of the steering wheel  \\
 & 3 & seat of the motorcycle & seat  \\
 & 4 & axle of the front wheel of the motorcycle & front wheel  \\
 & 5 & axle of the back wheel of the motorcycle & back wheel  \\
 & 6 & bottom part of the frame of the motorcycle & bottom of the engine  \\
 & 7 & axle of the front wheel of the motorcycle & front wheel  \\
 & 8 & axle of the back wheel of the motorcycle & back wheel  \\
 & 9 & front part of the front wheel of the motorcycle & front of the front wheel  \\
 & 10 & back part of the back wheel of the motorcycle & back of the back wheel  \\
 & 11 & top of the fuel tank of the motorcycle & top of the tank  \\
 & 12 & front end of the front fender of the motorcycle & top of the front wheel  \\
 & 13 & back end of the rear fender of the motorcycle & top of the back wheel  \\
\hline
\multirow{31}{*}{Airplane}  & 0 & nose of the airplane & nose of the airplane  \\
 & 1 & cockpit of the airplane & top of the cockpit  \\
 & 2 & tail of the airplane & back end  \\
 & 3 & the end of the left wing of the airplane & tip of all the wings  \\
 & 4 & wing root of the left wing of the airplane & different corners of the intersection between the body and the wings  \\
 & 5 & the end of the right wing of the airplane & tip of all the wings  \\
 & 6 & wing root of the right wing of the airplane & different corners of the intersection between the body and the wings  \\
 & 7 & wingtip of the left horizontal stabilizer of the airplane & tip of all the wings  \\
 & 8 & wingtip of the right horizontal stabilizer of the airplane & tip of all the wings  \\
 & 9 & top of vertical stabilizer of the airplane & - \\
 & 10 & wingtip of the left diagonal stabilizer of the airplane & tip of all the wings  \\
 & 11 & wingtip of the right diagonal stabilizer of the airplane & tip of all the wings  \\
 & 12 & wing root of the left wing of the airplane & different corners of the intersection between the body and the wings  \\
 & 13 & wing root of the right wing of the airplane & different corners of the intersection between the body and the wings  \\
 & 14 & inner turbine of the left wing of the airplane & center of all turbines  \\
 & 15 & inner turbine of the right wing of the airplane & center of all turbines  \\
 & 16 & outer turbine of the left wing of the airplane & center of all turbines  \\
 & 17 & outer turbine of the right wing of the airplane & center of all turbines  \\
 & 18 & outer turbine of the right wing of the airplane & - \\
\hline
\end{tabular}
\end{table*}

\begin{table*}
\centering
\begin{tabular}{|l|l|l|l|}
\hline
Object & Index & Description & Short Description \\
\hline
\multirow{21}{*}{Car} & 0 & front right wheel of the car & front right wheel  \\
 & 1 & rear right wheel of the car & back right wheel  \\
 & 2 & rear left wheel of the car & back left wheel  \\
 & 3 & front left wheel of the car & front left wheel  \\
 & 4 & corner of the front bumper of the car & right part of the front bumper  \\
 & 5 & corner of the rear bumper of the car & left part of the back bumper  \\
 & 6 & corner of the rear bumper of the car & left part of the back bumper  \\
 & 7 & corner of the front bumper of the car & right part of the front bumper  \\
 & 8 & top corner of the windshield of the car & top of right window  \\
 & 9 & top corner of the windshield of the car & top right corner of back windshield  \\
 & 10 & top corner of the rear window of the car & left corner of back windshield  \\
 & 11 & top corner of the rear window of the car & top left corner of front windshield  \\
 & 12 & bottom corner of the windshield of the car & bottom left corner of front windshield  \\
 & 13 & bottom corner of the windshield of the car & bottom right corner of front windshield  \\
 & 14 & bottom corner of the rear window of the car & bottom right corner of back windshield  \\
 & 15 & bottom corner of the rear window of the car & bottom left corner of back windshield  \\
 & 16 & right headlight of the car & front right headlight  \\
 & 17 & right taillight of the car & right back headlight  \\
 & 18 & left taillight of the car & left back headlight  \\
 & 19 & left headlight of the car & front left headlight  \\
 & 20 & right side mirror of the car & right rear view  \\
 & 21 & left side mirror of the car & left rear view  \\
\hline
\multirow{8}{*}{Guitar} & 0 & head of the guitar & top of the headstock  \\
 & 1 & connection of the body and neck of the guitar & middle of the fretboard  \\
 & 2 & corner of the body of the guitar & top left corner of the body  \\
 & 3 & corner of the body of the guitar & top right corner of the body  \\
 & 4 & corner of the body of the guitar & bottom left corner of the body  \\
 & 5 & corner of the body of the guitar & bottom right corner of the body  \\
 & 6 & strap pin on the bottom of the body of the guitar & bottom of the body  \\
 & 7 & waist of the body of the guitar & middle left of the body  \\
 & 8 & waist of the body of the guitar & middle right of the body  \\

\hline
\end{tabular}
\end{table*}

\begin{table*}
\centering
\begin{tabular}{|l|l|l|l|}
\hline
Object & Index & Description & Short Description \\
\hline
\multirow{23}{*}{Bathtub}  & 0 & top outer corner of the bathtub & - \\
 & 1 & top outer corner of the bathtub & - \\
 & 2 & top outer corner of the bathtub & - \\
 & 3 & top outer corner of the bathtub & - \\
 & 4 & bottom outer corner of the bathtub & - \\
 & 5 & bottom outer corner of the bathtub & - \\
 & 6 & bottom outer corner of the bathtub & - \\
 & 7 & bottom outer corner of the bathtub & - \\
 & 8 & top inner corner of the bathtub & - \\
 & 9 & top inner corner of the bathtub & - \\
 & 10 & top inner corner of the bathtub & - \\
 & 11 & top inner corner of the bathtub & - \\
 & 12 & bottom inner corner of the bathtub & - \\
 & 13 & bottom inner corner of the bathtub & - \\
 & 14 & bottom inner corner of the bathtub & - \\
 & 15 & bottom inner corner of the bathtub & - \\
 & 16 & center of the short side upper edge of the bathtub & - \\
 & 17 & center of the short side upper edge of the bathtub & - \\
 & 18 & center of the short side lower edge of the bathtub & - \\
 & 19 & center of the short side lower edge of the bathtub & - \\
 & 20 & center of the short side upper edge of the bathtub & - \\
 & 21 & center of the short side upper edge of the bathtub & - \\
 & 22 & center of the short side lower edge of the bathtub & - \\
 & 23 & center of the short side lower edge of the bathtub & - \\
\hline
\multirow{8}{*}{Helmet} & 0 & top of the helmet & top  \\
 & 1 & front of the helmet peak & front  \\
 & 2 & back of the helmet peak & back  \\
 & 3 & left side of the helmet peak & left  \\
 & 4 & right side of the helmet peak & right  \\
 & 5 & right bottom of the helmet & bottom right  \\
 & 6 & left bottom of the helmet & bottom left  \\
 & 7 & bottom of the visor of the helmet & - \\
 & 8 & bottom of the back of the helmet & bottom back  \\
\hline
\end{tabular}
\end{table*}

In this supplementary material, we present a comprehensive text prompt outlining key points for 16 common object categories used in KeypointNet. Each object category is defined by a specific set of semantically meaningful manually annotated key points that capture its structural and functional characteristics. These key points are carefully selected to represent distinctive geometric features and important structural elements of each object. However, in KeypointNet, these key points are not accompanied by a text description; instead, they are assigned a unique semantic index. Some of these labels are derived from the source referenced in \cite{wimmer2024back}. This information may also be useful for others aiming to detect key points based solely on textual descriptions.

\end{document}